\documentclass{article}
\usepackage{log_2026}						

\usepackage{booktabs}						
\usepackage{multirow}						
\usepackage{amsfonts}						
\usepackage{graphicx}						
\usepackage{duckuments}						
\usepackage{float}
\usepackage{longtable}
\usepackage{booktabs}
\usepackage{array}
\usepackage[numbers,compress,sort]{natbib}	

\title[Support-Set Target Leakage in RFMs during ICL]{Support-Set Target Leakage in Relational Foundation Models during In-Context Learning: Impact, Detection, and Mitigation}

\author[Prefect et al.]{%
Roshan Reddy Upendra \\
SAP \\
\email{roshan.reddy.upendra@sap.com}\And
Alexandre Dorais \\
SAP \\
\email{alexandre.dorais@sap.com}\And
Joe Meyer \\
SAP \\
\email{joseph.meyer@sap.com}\And
Andrew Pouret \\
SAP \\
\email{andrew.pouret@sap.com}\And
Anastasios Lambrianos Stappas \\
SAP \\
\email{anastasios.lambrianos.stappas@sap.com}\And
Dinesh Katupputhur Ramprasath \\
SAP \\
\email{dinesh.katupputhur.ramprasath@sap.com}\And
Tom Palczewski \\
SAP \\
\email{tom.palczewski@sap.com}\And
Minghua Li \\
SAP \\
\email{mark.li01@sap.com}
}

\begin{document}

\maketitle

\begin{abstract}
Relational in-context learning (ICL) conditions predictions on the labeled support examples and their linked tables, creating a failure mode when the support set contains target-derived features that are unavailable for the query. We formulate this problem as support-set target leakage, distinct from leakage during dataset construction, temporal splitting, or representation learning. Here, the target-derived (leaker) columns are present only in the labeled support set during relational in-context inference, while queries remain clean. We construct 14 synthetic leaker types, corresponding to 20 columns, spanning proxies with different noise levels, coverage, modalities, semantic transparency, and relational distances. We evaluate a frozen relational encoder with an ICL head on held-out RelBench databases and use Integrated Gradients (IG) to rank and remove suspicious columns. Our results show that the effect of support-set leakage varies across tasks and relational distances. Target-table leakers cause the clearest degradation, while one- and two-hop leakers are not consistently used by the model. IG ranks target-table leakers highly across datasets and partially recovers performance in settings where leakage has the largest effect.
\end{abstract}

\section{Introduction}

Relational foundation models (RFMs) transfer representations across databases, schemas, and tasks. Recent approaches include the zero-shot Relational Transformer \cite{ranjan2025relational}, the graph-centric Griffin encoder \cite{wang2025griffin}, and relational in-context learning (ICL) models such as KumoRFM-2 \cite{hudovernik2026kumorfm}, RDB-PFN \cite{wang2026relational}, and OpenRFM \cite{chen2026openrfm}. Meyer et al. \cite{meyer2026relational} similarly combine relational neighborhood aggregation with a tabular foundation-model head for training-free ICL.

Relational databases distribute information across linked tables, so target-derived or post-outcome attributes may appear several joins from the target. Kaufman et al.~\cite{kaufman2012leakage} describe a customer-prediction system in which archived company websites contained references to products purchased only after the prediction date. These references were predictive in historical data but unavailable for new customers, and temporal filtering could not remove them when earlier website versions were unavailable. A similar mismatch can occur in relational ICL when support examples or their linked records contain target-derived information absent from the query. 

Related work studies adjacent failure modes. ICLShield \cite{ren2025iclshield} examines backdoor attacks in language-model ICL, where poisoned demonstrations induce trigger-based predictions. OpenRFM \cite{chen2026openrfm} perturbs labels and features to test relational-context use. GRAFT \cite{sahoo2026graft} applies Integrated Gradients (IG) \cite{sundararajan2017axiomatic} to detect injected spurious features in GNN classifiers, while recent TabPFN work \cite{hu2026noise} studies robustness to irrelevant and correlated features and label noise. Task Scarcity and Label Leakage \cite{azevedo2026task} studies task-specific shortcuts learned during relational transfer and removes label-predictive directions during training. In contrast, we study target-derived relational columns that are present only in the ICL support set and unavailable to the clean query.

Detecting such columns is difficult because target association alone does not imply leakage, while true leakers may be noisy, sparse, opaque, or relationally distant. Provenance, timestamps, domain knowledge, correlation, and mutual information can identify suspicious columns but do not establish model use \cite{kaufman2012leakage}. Leave-one-column-out and permutation importance probe the downstream model more directly, but redundant leakers can mask one another and each candidate requires separate evaluation \cite{breiman2001random,strobl2008conditional}. Model-agnostic SHAP considers feature subsets but is computationally costly and depends on how missing and correlated features are represented \cite{lundberg2017unified,aas2021explaining}. Attention and intermediate representations may indicate where information is encoded without showing whether it determines the final prediction \cite{jain2019attention,wiegreffe2019attention,ravichander2021probing}. These limitations motivate IG as a joint screening method.

To the best of our knowledge, support-set target leakage in relational ICL has not been systematically studied. This problem is particularly relevant to enterprise databases, where features may be drawn from several related tables and may not all be available at prediction time. We study how target-derived columns in the support set affect predictions on clean queries. Using a pretrained relational encoder, we introduce synthetic leakers only during downstream inference and compare clean-support and leaky-support conditions while keeping the query unchanged. We then use IG to rank influential relational columns, remove the flagged columns, and measure the resulting change in clean-query performance. Our experiments examine both direct target-table leakage and leakers placed one or two relational hops away. Figure \ref{fig:pipeline} in Appendix \ref{app:protocol} summarizes the clean, leaky, and detection/mitigation evaluation pipeline. Our contributions are:
\begin{enumerate}
\item A controlled formulation of support-set target leakage in relational ICL, where target-derived columns are available only in the support set and absent from the clean query set.
\item A 20-column stress test spanning different proxy types, modalities, semantic transparency, and relational distances, together with matched zero-, one-, and two-hop ablations.
\item An initial attribution-based approach that does not use known leaker identities, for ranking suspicious relational columns and evaluating the effect of their removal on clean-query performance.
\end{enumerate}

\section{Methodology}

\subsection{Problem Formulation}

Let $\mathcal{D}$ and $\mathcal{G}(\mathcal{D})= (\mathcal{V},\mathcal{E})$ denote a relational database and its heterogeneous graph representation, respectively, where each database row is represented as a node and primary-foreign key relationships define edges. For a downstream classification task, the model receives a labeled support set \(\mathcal{S}_{\mathrm{clean}}=\{(x_i,y_i)\}_{i=1}^{n}\) and predicts labels for a query set \(\mathcal{Q}_{\mathrm{clean}}=\{x_j\}_{j=1}^{m}\). Here, $x_i$ denotes the target row together with the relational context sampled around it by the encoder.

We study a setting in which one or more columns in the support context contain target-derived information. For a leaker column $c$ stored on a row $v$, its value is generated as \(z_{v,c}=g_c\!\left(\{y_k:k\in\mathcal{A}(v)\}\right)\), where $\mathcal{A}(v)$ denotes the labeled target rows associated with $v$. For a leaker stored directly on target row $i$, this reduces to $z_{i,c}=g_c(y_i)$. The function $g_c$ may produce a deterministic, noisy, partially observed, numerical, categorical, or text-valued proxy.

Let $\widetilde{x}_i$ denote the relational context of support example $i$ after the leaker columns have been introduced. The contaminated support set is \(\widetilde{\mathcal{S}}_{\mathrm{leaky}}=\{(\widetilde{x}_i,y_i)\}_{i=1}^{n}\). The corresponding leaker values are unavailable in the query contexts. We therefore compare \(f_{\theta}(\mathcal{S}_{\mathrm{clean}},\mathcal{Q}_{\mathrm{clean}}) \quad\text{and}\quad f_{\theta}(\widetilde{\mathcal{S}}_{\mathrm{leaky}},\mathcal{Q}_{\mathrm{clean}})\), where the parameters $\theta$ are not updated during downstream inference. This comparison isolates the effect of information available only in the support context. Ground-truth leaker identities are used to construct the controlled mismatch and evaluate detection, but are not provided to the detector.

\subsection{Synthetic Leaker Construction}

We inject 20 target-derived columns that vary in signal fidelity, coverage, data type, semantic transparency, and relational distance. The full stress test includes 14 target-table leakers, three one-hop leakers, and three two-hop leakers jointly, where interactions between columns may also occur. To study the effect of relational distance separately, we also evaluate the same three proxy types in zero-hop-only, one-hop-only, and two-hop-only settings. The matched hop ablations use the same three proxy designs at each distance to vary relational placement while keeping their construction as comparable as possible. Full construction details are provided in Appendix \ref{app:leaker_catalog}. 

\subsection{Relational ICL Pipeline}

We construct sampled relational subgraphs around each target row and encode them with a frozen pretrained Griffin encoder \cite{wang2025griffin}. The 512-dimensional target-row representation after two relational message-passing layers is used as the embedding. Full encoder settings are in Appendix \ref{app:model_details}. For each task, the training and validation embeddings and labels form the support set for the pretrained TabPFN-based ICL head \cite{hollmann2022tabpfn}, while test embeddings form the query set. Each ablation includes only the selected leaker columns in the support data, with the query kept clean.

\subsection{Attribution-Based Leaker Detection}
\label{LeakerDetection}

We perform leaker-annotation-free detection by scoring all relational columns. On the validation set, we compute IG through the frozen Griffin encoder using the Ridge classifier as a surrogate. IG measures the contribution of each representation by integrating gradients of the surrogate output along a straight-line path from a zero baseline to the observed representation. We use Ridge instead of the full TabPFN ICL head because it provides a lightweight differentiable objective for IG over Griffin embeddings. The resulting scores identify columns that are influential for the surrogate and need not exactly match the columns used by downstream ICL. We therefore interpret IG as a screening method rather than a faithful explanation of the TabPFN head. Mitigation is evaluated separately by recomputing the Griffin embeddings and measuring TabPFN head performance after column removal, without assuming attribution equivalence between the prediction heads. Candidate columns are scored jointly at each IG step, avoiding a separate model evaluation for every column.

We combine each column's mean attribution with the mean attribution among its most influential occurrences. The resulting scores are normalized within groups defined by source table and feature type. Columns with unusually large normalized scores are flagged using a Bonferroni-style cutoff applied to a one-sided normal-tail transform of the robust scores. This cutoff is used as a heuristic screening rule rather than a calibrated statistical test. Full details are in Appendix \ref{app:attribution_thresholding}.

\subsection{Mitigation and Evaluation}

To evaluate mitigation, we remove all flagged columns from both the support and query set, including their values and feature-name representations. We then recompute the Griffin embeddings and repeat TabPFN inference. This removal uses the detector output directly and hence includes both true and false positive columns. We evaluate a clean baseline, a full stress test containing all 20 leaker columns, and matched zero-hop-only, one-hop-only, and two-hop-only ablations containing three leakers each. IG is computed on the validation split, and columns are flagged using the heuristic cutoff described in Section \ref{LeakerDetection}. We compare three settings: a clean baseline with no injected leakers, leaky support with a clean query, and leaky support post-removal of flagged columns from both the support and query sets. We treat this as an initial proof-of-concept study to establish the impact of support-set leakage in relational ICL rather than a final automatic mitigation procedure. 

\section{Experimental Setup}

The pretrained Griffin encoder was trained on relational data from 15 CTU datasets \cite{motl2015ctu} and four RelBench databases \cite{robinson2024relbench}. We evaluate downstream relational ICL on five classification tasks from three held-out RelBench databases: \texttt{rel-avito}, \texttt{rel-event}, and \texttt{rel-trial}. The complete pretraining dataset list is provided in Appendix \ref{app:pretraining_data}. For each task, the training and validation splits form the labeled support set, and the test split is used as the query set. The Griffin encoder remains frozen, and a pretrained TabPFN classifier is fitted on the support embeddings for each evaluation setting. All paired runs use the same random seed. An additional Bimodal encoder robustness check over 13 tasks is reported in Appendix \ref{app:bimodal}. Other implementation details are in Appendix \ref{app:protocol}.

\section{Results and Discussion}

Table \ref{tab:combined_results} shows that the effect of support-only leakage varies across datasets. Matched zero-hop leakers caused the largest decreases in \texttt{rel-event} and \texttt{rel-trial}, while \texttt{rel-avito} was unchanged. The two \texttt{rel-avito} tasks have low clean macro-F1 and remain unchanged across all leakage ablations. Therefore, we do not interpret these results as evidence of robustness to leakage. Although \texttt{rel-trial} also has low clean macro-F1, matched zero-hop leakage decreases performance from 0.500 to 0.435, showing a measurable effect under the controlled support-only mismatch. The full stress test substantially reduced \texttt{rel-event} performance, whereas the higher-hop ablations produced small changes. Thus, support-only target-derived columns affect clean-query predictions only when the model relies on their signal. An additional Bimodal encoder check also yields a larger mean effect at zero-hop than at one- or two-hops across 13 tasks (mean $\Delta F1=-0.115,-0.002,+0.001$, respectively), although task-level effects vary (Appendix \ref{app:bimodal}). Per-leaker ablations show that the zero-hop effect spans multiple proxy types, while matched higher-hop effects remain small in this pipeline (Appendix \ref{app:leaker_type_effects}). Within the tested designs, relational placement gives the clearer pattern, i.e., the matched leakers have their largest effects at zero-hop, while semantically transparent text proxies are more harmful on average at zero-hop than the deterministic numerical proxy, suggesting semantic cues can influence leakage effects, although differences in encoding and column-name explicitness prevent isolating the role of semantic cues.

Table \ref{tab:combined_results} also summarizes leaker-detection performance across the same ablations. Zero-hop AUROC exceeded $0.9$ on all five tasks across the three evaluated datasets. However, no column crossed the heuristic cutoff on \texttt{rel-avito}, giving zero recall and detection F1. This does not imply random ranking; rather, the leakers were ranked above many non-leakers but were not sufficiently separated to pass the threshold. The decision-tree sanity check (Appendix \ref{app:leaker_type_effects}) confirms that the same \texttt{rel-avito} leakers are predictive when available in both training and test data, while the support-only Griffin-TabPFN runs remain unchanged. This rules out missing target signal in the injected columns, but does not determine whether the signal is attenuated by the encoder, relational aggregation, or the ICL head. Since clean and leaky macro-F1 were also unchanged on \texttt{rel-avito}, the current results do not show that these columns materially affected its predictions. All candidate columns are attributed in the same IG passes, avoiding a separate encoder evaluation for each column. However, when several leakers provide similar target-derived signals, their attribution may be divided across columns, causing individual leakers to remain below the detection threshold. On \texttt{rel-trial}, this may explain why one highly ranked zero-hop proxy remained unflagged. Group-aware thresholding could address this without changing the joint IG computation.

\begin{table}[t]
\centering
\caption{
Dataset-level results across leakage ablations. ICL macro-F1 is reported for  clean support $\mathbf{F1_{\mathrm{clean}}}$, leaky support $\mathbf{F1_{\mathrm{leaky}}}$, and post-detection removal $\mathbf{F1_{\mathrm{flagged}}}$. Leaker detector F1 is reported as $\mathbf{F1_{\mathrm{detect}}}$. Values are averaged across tasks within each dataset. Per-task results in Appendix \ref{app:per_task_results}.
}
\label{tab:combined_results}
\small
\resizebox{\textwidth}{!}{
\begin{tabular}{llccccc@{\hspace{8pt}}cccc}
\toprule
\textbf{Dataset}
& \textbf{Leakage setting}
& \textbf{\# Tasks}
& \textbf{\# Leakers}
& \multicolumn{3}{c}{\textbf{ICL inference}}
& \multicolumn{4}{c}{\textbf{Leaker detector}} \\
\cmidrule(lr){5-7}
\cmidrule(lr){8-11}
&
&
&
&
$\mathbf{F1_{\mathrm{clean}}}$
& $\mathbf{F1_{\mathrm{leaky}}}$
& $\mathbf{F1_{\mathrm{flagged}}}$
& \textbf{Precision}
& \textbf{Recall}
& $\mathbf{F1_{\mathrm{detect}}}$
& \textbf{AUROC} \\
\midrule

& Full stress test
& & 20
& & 0.517 & 0.546
& 0.404 & 0.475 & 0.437 & 0.937 \\

rel-event
& Leakers, zero-hop only
& 2 & 3
& 0.674 & 0.500 & 0.589
& 0.158 & 1.000 & 0.273 & 0.993 \\

& Leakers, one-hop only
& & 3
& & 0.688 & 0.561
& 0.154 & 1.000 & 0.267 & 0.880 \\

& Leakers, two-hop only
& & 3
& & 0.691 & 0.528
& 0.029 & 0.167 & 0.050 & 0.653 \\

\midrule

& Full stress test
& & 20
& & 0.498 & 0.443
& 1.000 & 0.250 & 0.400 & 0.906 \\

rel-trial
& Leakers, zero-hop only
& 1 & 3
& 0.501 & 0.435 & 0.434
& 1.000 & 0.667 & 0.800 & 0.991 \\

& Leakers, one-hop only
& & 3
& & 0.542 & 0.525
& 0.000 & 0.000 & 0.000 & 0.986 \\

& Leakers, two-hop only
& & 3
& & 0.484 & 0.490
& 0.000 & 0.000 & 0.000 & 0.430 \\

\midrule

& Full stress test
& & 20
& & 0.478 & 0.478
& 0.000 & 0.000 & 0.000 & 0.815 \\

rel-avito
& Leakers, zero-hop only
& 2 & 3
& 0.478 & 0.478 & 0.478
& 0.000 & 0.000 & 0.000 & 0.911 \\

& Leakers, one-hop only
& & 3
& & 0.478 & 0.478
& 0.000 & 0.000 & 0.000 & 0.706 \\

& Leakers, two-hop only
& & 3
& & 0.478 & 0.478
& 0.000 & 0.000 & 0.000 & 0.411 \\

\bottomrule
\end{tabular}
}
\end{table}

Post-detection removal recovered part of the largest \texttt{rel-event} losses in the full stress test. On \texttt{rel-trial} zero-hop, two of the three leakers were removed with no false positives, but the remaining deterministic proxy was ranked sixth and did not pass the threshold; macro-F1 therefore remained near the leaky result. In the full stress test, $15$ of the $20$ leakers remained after removal, so the results do not isolate the effect of the detected subset. For the \texttt{rel-event} one- and two-hop settings, leakage caused no degradation, while removal included many legitimate columns and reduced performance. These cases show that useful attribution rankings are not sufficient for safe mitigation: missed leakers can preserve the mismatch, while false positives can remove predictive information. 

\section{Conclusion}

We present a controlled study of support-set target leakage in relational ICL and show that target-derived features available only in the support set can materially affect predictions on clean queries. Across five held-out tasks, zero-hop leakers produced the clearest performance losses, indicating that relational distance matters in our Griffin-TabPFN pipeline. IG consistently ranked zero-hop leakers highly across datasets, while thresholded removal remained sensitive to calibration, false positives, and missed leakers. These results establish support-set-only target-derived features as a practical concern for relational ICL and motivate developing reliable leakage detection and mitigation methods. Limitations and future work are discussed in Appendix \ref{app:limitations}.

\bibliographystyle{unsrtnat}
\bibliography{reference}

\appendix

\section{Experimental Protocol}
\label{app:protocol}

The evaluation tasks and relational placements are summarized in Table \ref{tab:appendix_task_setup}.  For every task, the original training and validation splits form the labeled support set, while the test split is used as the clean query set. Synthetic leaker columns are injected only into the support data. Before computing query embeddings, all injected columns are physically removed from the query schema, including both their value representations and feature-name embeddings. In each ablation, all injected columns not belonging to that setting are likewise physically removed from the support schema. The pretrained Griffin encoder remains frozen throughout evaluation.

IG is computed on the validation portion of the support set. After detection, every flagged column, including both true and false positives, is physically removed from the support and query schemas. Griffin embeddings are then recomputed from the modified data before repeating TabPFN inference. For tasks with timestamps, support examples and sampled relational neighborhoods are restricted to information available at or before the corresponding query time.

\begin{figure}[t]
\centering
\includegraphics[width=1.0\linewidth]{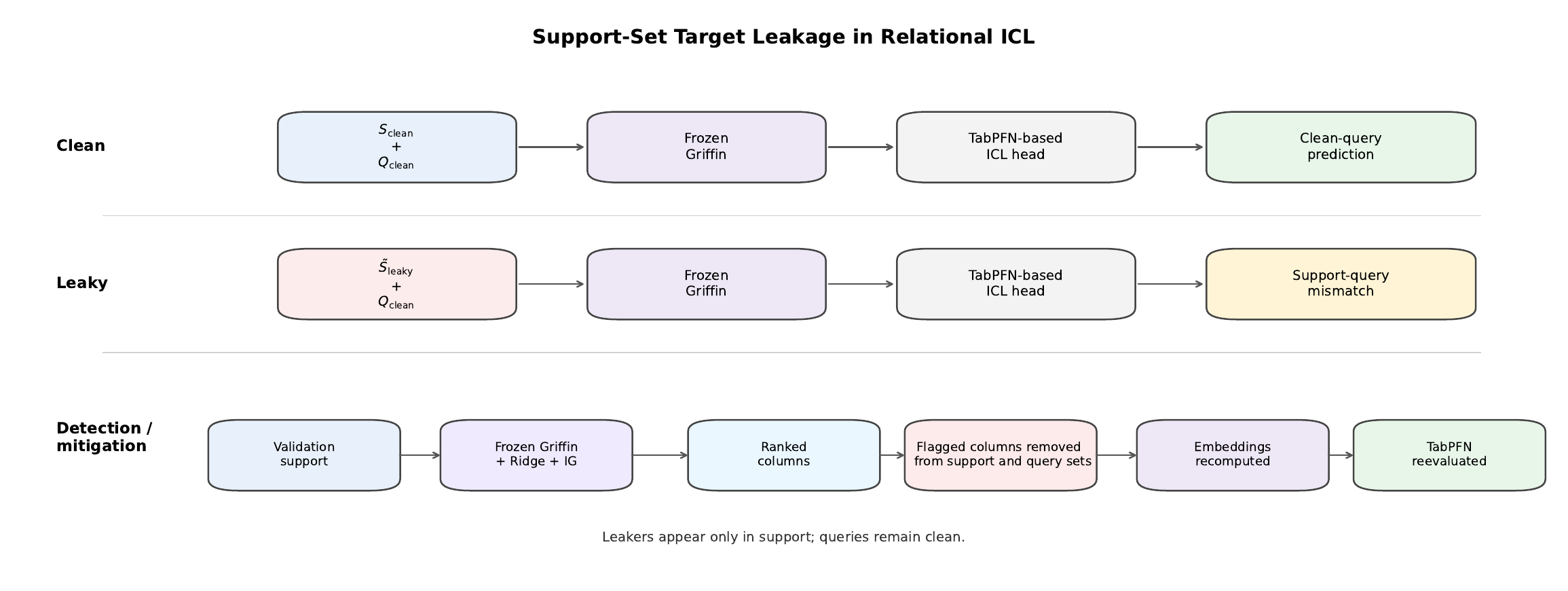} \caption{Support-set target leakage pipeline. Clean and leaky support settings are evaluated with Griffin and TabPFN; IG with a Ridge surrogate ranks columns for removal and reevaluation.}
\label{fig:pipeline}
\end{figure}

\begin{table}[H]
\centering
\caption{Evaluation tasks and relational placement of the injected columns. Candidate columns are reported for the clean baseline, the matched three-column hop ablations, and the full 20-column stress test, respectively. All evaluated tasks are binary classification tasks.}
\label{tab:appendix_task_setup}
\small
\resizebox{\textwidth}{!}{
\begin{tabular}{llrrrrlll}
\toprule
\textbf{Dataset} &
\textbf{Task} &
\textbf{Support} &
\textbf{Query} &
\textbf{\# Classes} &
\textbf{Candidates} &
\textbf{Target table} &
\textbf{1-hop table} &
\textbf{2-hop table} \\
\midrule
\texttt{rel-avito}
& \texttt{user-clicks}
& 80,637
& 47,996
& 2
& 30 / 33 / 50
& \texttt{UserInfo}
& \texttt{VisitStream}
& \texttt{AdsInfo} \\

\texttt{rel-avito}
& \texttt{user-visits}
& 116,598
& 36,129
& 2
& 30 / 33 / 50
& \texttt{UserInfo}
& \texttt{VisitStream}
& \texttt{AdsInfo} \\

\texttt{rel-event}
& \texttt{user-ignore}
& 23,424
& 3,949
& 2
& 125 / 128 / 145
& \texttt{users}
& \texttt{event\_interest}
& \texttt{events} \\

\texttt{rel-event}
& \texttt{user-repeat}
& 4,110
& 246
& 2
& 125 / 128 / 145
& \texttt{users}
& \texttt{event\_interest}
& \texttt{events} \\

\texttt{rel-trial}
& \texttt{study-outcome}
& 12,954
& 825
& 2
& 117 / 120 / 137
& \texttt{studies}
& \texttt{conditions\_studies}
& \texttt{conditions} \\
\bottomrule
\end{tabular}
}
\end{table}

\section{Pretraining Data, Model, and Checkpoint Details}
\label{app:model_details}

\subsection{Pretraining and Held-Out Databases}
\label{app:pretraining_data}

The frozen Griffin encoder was pretrained on clean relational data from 15 CTU databases and four RelBench databases. The complete pretraining set is listed in Table  \ref{tab:pretraining_datasets}. The downstream evaluation databases \texttt{rel-avito}, \texttt{rel-event}, and \texttt{rel-trial} were excluded from pretraining and were used only for the relational in-context-learning experiments reported in this work. 

\begin{table}[H]
\centering
\caption{Databases used to pretrain the Griffin checkpoint. The three
evaluation databases were held out from pretraining.}
\label{tab:pretraining_datasets}
\small
\begin{tabular}{lp{0.78\textwidth}}
\toprule
\textbf{Source} & \textbf{Databases} \\
\midrule
CTU &
\texttt{ctu-accidents},
\texttt{ctu-airline},
\texttt{ctu-chess},
\texttt{ctu-dallas},
\texttt{ctu-ergastf1},
\texttt{ctu-financial},
\texttt{ctu-ftp},
\texttt{ctu-geneea},
\texttt{ctu-legalacts},
\texttt{ctu-mondial},
\texttt{ctu-ncaa},
\texttt{ctu-premiereleague},
\texttt{ctu-thrombosis},
\texttt{ctu-tpcc}, and
\texttt{ctu-voc} \\

RelBench &
\texttt{rel-amazon},
\texttt{rel-f1},
\texttt{rel-hm}, and
\texttt{rel-stack} \\

\bottomrule
\end{tabular}
\end{table}

\subsection{Model and Inference Configuration}
\label{app:model_details}

We use the same frozen Griffin checkpoint for the clean, leaky, and detected-removal conditions. The encoder uses 512-dimensional representations, two relational message-passing layers, eight attention heads, dropout 0.1, and a neighborhood fanout of 9. Temporal filtering is enabled. Griffin is kept frozen during all downstream experiments, and embeddings are recomputed whenever injected or detected columns are removed.

The resulting support and query embeddings are passed to \texttt{TabPFNClassifier} using the checkpoint \texttt{tabpfn-v3-classifier-v3\_default.ckpt} \cite{grinsztajn2026tabpfn} and with 32 estimators. The original training and validation embeddings form the labeled support set, while test embeddings form the clean query set. We use the same Griffin and TabPFN configurations across clean, leaky, and post-removal conditions.

\section{Synthetic-Leaker Construction}
\label{app:leaker_catalog}

The full stress test contains 20 injected columns: 14 on the task target table, three on a directly related one-hop table, and three on a selected two-hop table. Table \ref{tab:leaker_catalog} summarizes their signal fidelity, representation, coverage, and semantic transparency. The physical source tables used for each task are reported in Table \ref{tab:appendix_task_setup}.

For numerical leakers, rows without an available support label are assigned the sentinel value $-1$. For text leakers, such rows are mapped to a dedicated null token or null phrase. Noisy proxies retain the true class with probabilities 0.80, 0.90, 0.70, or 0.50, and otherwise, the class is replaced by a different class, which is equivalent to flipping the label for the binary tasks evaluated here. The partial-coverage proxy is generated for an independently sampled 50\% of labeled rows and is set to $-1$ elsewhere.

Opaque categorical values are deterministic 16-character tokens obtained from SHA-256 hashes of the dataset, leaker identifier, and class. Thus, each class is represented consistently within an experiment, but the token itself contains no class semantics. Semantic text values are task-specific natural-language class paraphrases encoded using the same text-embedding pipeline as the original relational text features.

At one hop, the label of each support target is copied to its directly associated row in the selected neighboring table. At two hops, labels from all support targets connected to a row through the selected two-edge path are collected and mean-aggregated. The numerical two-hop proxy stores this mean directly. For the text variants, the mean is rounded to the nearest
class before selecting the corresponding opaque token or semantic phrase. Rows that are not connected to any labeled support target receive the numerical or textual null representation.

To isolate the effect of relational distance, we instantiate the same three complementary proxy designs at zero, one, and two hops: (i) a deterministic numerical proxy with a non-semantic column name, (ii) a proxy with a target-informative column name and opaque class-dependent values, and (iii) a proxy with a non-semantic column name and semantically meaningful class-dependent values. These designs separately expose numerical target signal, column-name semantics, and encoded-value semantics while keeping the proxy construction as comparable as possible across relational distances.

Ground-truth leaker identities are used only to construct the controlled ablations and calculate detection metrics. They are not supplied to the leaker detector or used when selecting columns for removal.

\small
\setlength{\tabcolsep}{3pt}
\renewcommand{\arraystretch}{1.08}

\begin{longtable}{@{}
p{0.30\textwidth}
p{0.07\textwidth}
p{0.12\textwidth}
p{0.21\textwidth}
p{0.21\textwidth}
@{}}

\caption{Complete catalog of the 20 synthetic leaker columns. \emph{Column-name semantics} indicates whether the column name explicitly reveals its relationship to the prediction target. \emph{Encoded-value semantics} indicates whether the stored values have human-interpretable target-class meaning.}
\label{tab:leaker_catalog}\\

\toprule
\textbf{Leaker type} &
\textbf{Hop} &
\textbf{Encoding} &
\textbf{Column-name semantics} &
\textbf{Encoded-value semantics} \\
\midrule
\endfirsthead

\multicolumn{5}{c}%
{{\tablename\ \thetable{} -- continued from previous page}}\\
\toprule
\textbf{Leaker type} &
\textbf{Hop} &
\textbf{Encoding} &
\textbf{Column-name semantics} &
\textbf{Encoded-value semantics} \\
\midrule
\endhead

\midrule
\multicolumn{5}{r}{{Continued on next page}}\\
\endfoot

\bottomrule
\endlastfoot

Perfect deterministic &
0 &
Numerical &
No &
No \\

Noisy 80\% &
0 &
Numerical &
No &
No \\

Rank/ordinal &
0 &
Numerical &
No &
No \\

Jittered numerical proxy &
0 &
Numerical &
No &
No \\

Semantic name, opaque values &
0 &
Text &
Yes &
No \\

Non-semantic name, semantic values &
0 &
Text &
No &
Yes \\

Semantic name and semantic values &
0 &
Text &
Yes &
Yes \\

Binned numerical proxy &
0 &
Numerical &
No &
No \\

Coarse categorical proxy &
0 &
Text &
No &
No \\

Noisy 90\% &
0 &
Numerical &
No &
No \\

Noisy 70\% &
0 &
Numerical &
No &
No \\

Noisy 50\% &
0 &
Numerical &
No &
No \\

Partial-coverage proxy &
0 &
Numerical &
No &
No \\

Opaque name and opaque values &
0 &
Text &
No &
No \\

Perfect deterministic &
1 &
Numerical &
No &
No \\

Semantic name, opaque values &
1 &
Text &
Yes &
No \\

Non-semantic name, semantic values &
1 &
Text &
No &
Yes \\

Deterministic aggregated proxy &
2 &
Numerical &
No &
No \\

Semantic name, opaque values &
2 &
Text &
Yes &
No \\

Non-semantic name, semantic values &
2 &
Text &
No &
Yes \\

\end{longtable}

\section{Leaker-Type and Relational-Distance Effects}
\label{app:leaker_type_effects}

We first use shallow decision trees to verify that the injected columns contain target signal. In this analysis, each leaker is available during both training and testing, so it is only a sanity check and does not model the support-only mismatch studied in the main experiments. All leakers together and each of the three matched zero-hop designs achieve macro-F1 of
1.0 across the five tasks. Several one- and two-hop leakers are also predictive, although their strength varies by task.

We then evaluate each leaker separately with the Griffin-TabPFN pipeline. The support set contains all original columns and exactly one injected leaker, while the query contains only the original columns. We report
\[
\Delta F1 =
F1_{\mathrm{one\text{-}leaker}} - F1_{\mathrm{clean}},
\]
using the clean baseline from the same evaluation run.

\begin{figure}[H]
\centering
\includegraphics[width=\textwidth]{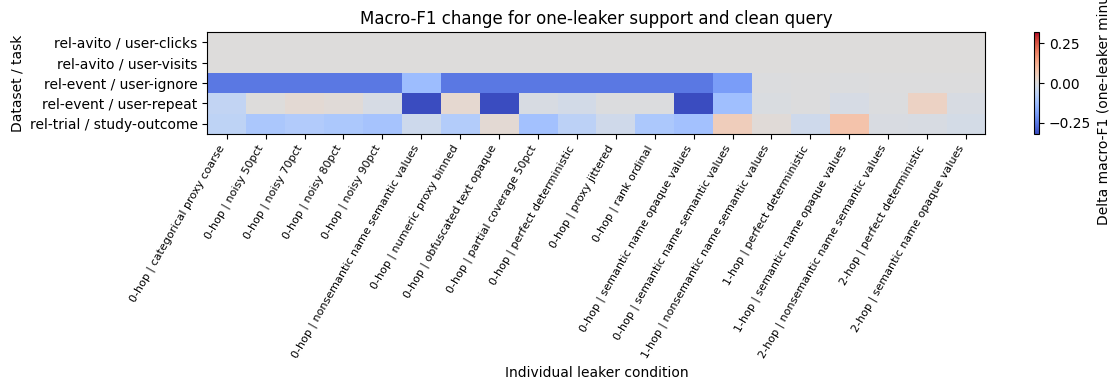} \caption{Per-leaker change in clean-query macro-F1 for all leaker types. Rows denote tasks and columns denote individual leaker types grouped by relational hop. Negative values indicate degradation relative to the corresponding clean-support baseline.}
\label{fig:leaker_effect_heatmap}
\end{figure}

Across the 70 zero-hop task-leaker conditions, the mean \(\Delta F1\) is \(-0.078\). Macro-F1 decreases in 36 conditions, increases in six, and is unchanged in 28. All unchanged cases are from the two \texttt{rel-avito} tasks. For the matched higher-hop leakers, the mean change is \(+0.001\) at one hop and \(-0.002\) at two hops. The zero-hop effect is therefore seen across several leaker types rather than being caused by one specific construction. Figure  \ref{fig:leaker_effect_heatmap} reports the per-leaker effects, including all 14 zero-hop leakers.

The 11 unmatched zero-hop leakers differ in noise, coverage, discretization, proxy fidelity, and whether their names or values carry semantic information. Excluding the two \texttt{rel-avito} tasks, they reduce macro-F1 in 27 of 33 conditions, with a mean \(\Delta F1\) of \(-0.120\). Partial coverage, the 90\%-fidelity proxy, the coarse categorical proxy, the rank/ordinal proxy, and the jittered proxy reduce F1 on all three remaining tasks. This shows that a support-only effect does not require an exact label copy or full coverage.

The effect of the other leakers varies by task. On \texttt{user-ignore}, every zero-hop leaker reduces F1, and most cause a similarly large drop. On \texttt{user-repeat}, the largest drops come from the semantic and opaque text proxies, while several noisy or binned numerical proxies have little effect or slightly increase F1. On \texttt{study-outcome}, noisy and partial-coverage numerical proxies are among the most harmful, while two text variants slightly increase F1. Overall, there is no consistent ordering across tasks by noise level, fidelity, or semantic transparency.

\begin{table}[H]
\centering
\caption{Per-leaker effects for the three matched proxy designs. Hop columns report the mean \(\Delta F1\) across the deterministic numerical, semantic-name with opaque-value, and non-semantic-name with semantic-value proxies.}
\label{tab:matched_one_leaker_effects}
\small
\begin{tabular}{llcccc}
\toprule
\textbf{Dataset} &
\textbf{Task} &
\(\mathbf{F1_{\mathrm{clean}}}\) &
\textbf{0-hop} &
\textbf{1-hop} &
\textbf{2-hop} \\
\midrule

\texttt{rel-avito} &
\texttt{user-clicks} &
0.481 &
0.000 &
0.000 &
0.000 \\

\texttt{rel-avito} &
\texttt{user-visits} &
0.475 &
0.000 &
0.000 &
0.000 \\

\texttt{rel-event} &
\texttt{user-ignore} &
0.721 &
-0.212 &
-0.003 &
-0.001 \\

\texttt{rel-event} &
\texttt{user-repeat} &
0.628 &
-0.222 &
-0.009 &
+0.008 \\

\texttt{rel-trial} &
\texttt{study-outcome} &
0.502 &
-0.075 &
+0.019 &
-0.016 \\

\bottomrule
\end{tabular}
\end{table}

Among the matched zero-hop designs, the semantic-name with opaque-value proxy has the largest average effect (\(\Delta F1=-0.138\)), followed by the non-semantic-name with semantic-value proxy (\(-0.097\)) and the deterministic numerical proxy
(\(-0.070\)). This ordering does not hold at one or two hops, where the effects are small and inconsistent. Since the designs also differ in encoding, these results do not isolate the effect of semantic transparency. 

The clearest pattern in Table \ref{tab:matched_one_leaker_effects} is the difference between zero-hop and higher-hop placement. All three matched designs have larger effects when placed on the target table. This suggests that leakage is weakened with relational distance in the current setup, but does not show that higher-hop leakage is generally harmless. Its effect may depend on connectivity, aggregation,
neighborhood sampling, and the encoder checkpoint.

These experiments also do not support a general claim that relational foundation models are robust or sensitive to leakage. They only show that the current frozen Griffin-TabPFN pipeline is sensitive to several target-table leakers and less affected by the tested one- and two-hop leakers.

\section{Per-Task Prediction and Detection Results}
\label{app:per_task_results}

Tables \ref{tab:task_prediction} and \ref{tab:task_detection} report results for every task and ablation. Clean denotes the baseline without injected leakers, Leaky denotes inference before removal, and Flagged denotes inference after all IG-flagged columns are physically removed from the support and query schemas and Griffin embeddings are recomputed. Performance entries are reported as macro-F1. All \(\Delta\) columns in Table \ref{tab:task_prediction} refer to macro-F1.

On the two \texttt{rel-avito} tasks, the injected leakers do not change prediction, so their target signal is not used by the current Griffin-TabPFN pipeline. On \texttt{rel-event}'s \texttt{user-ignore} task, zero-hop leakage is strongly harmful, but recovery is limited by the removal of many original columns. On \texttt{rel-event}'s \texttt{user-repeat} task, removal recovers more of the loss, although false positives still prevent consistent recovery. The higher-hop runs further show that removing flagged columns can reduce performance even when the injected leakers had little effect. On \texttt{rel-trial}'s \texttt{study-outcome} task, IG flags the two text leakers but misses the numerical proxy because its score does not pass the heuristic cutoff under the Ridge surrogate. Removing all three leakers in the oracle run restores performance to the clean baseline, showing that the missed proxy, or an interaction involving it, remains important. A practical next step is therefore to validate the highest-ranked unflagged columns with direct Griffin-TabPFN removal tests rather than lowering the threshold for all columns.

The detection results explain the mitigation behavior above. On the two \texttt{rel-avito} tasks, the matched zero-hop leakers are ranked above many original columns (\(\mathrm{AUROC}=0.911\)) but none pass the heuristic cutoff. Since these leakers also do not change prediction, the current results do not support lowering the threshold for these tasks. On \texttt{rel-event}, all three matched zero-hop leakers are detected, but 16 original columns are also flagged. The same threshold flags 17 columns in the clean runs. Thus, the limited recovery is mainly associated with low precision: removing useful original features offsets the benefit of removing the leakers. In the full stress test, several leakers are also missed, which further limits recovery. On \texttt{rel-trial}, ranking is strong, but thresholded recall is limited. Two of the three zero-hop leakers are flagged with no false positives, while the numerical proxy remains below the threshold. This is consistent with the oracle result, where removing all three restores  performance close to clean. The current scores are therefore useful for ranking, but high-ranked unflagged columns may require a second-stage Griffin--TabPFN removal test before mitigation.

Overall, the task-level results separate ranking, thresholded detection, and post-removal recovery: IG can rank leakers well even when none pass the threshold, while thresholding can still flag many original columns, so recovery depends on removing harmful leakers without removing useful features or missing important ones. A single global removal rule is therefore insufficient: lowering the threshold is not justified on \texttt{rel-avito}, \texttt{rel-event} needs better calibration to reduce false positives, and \texttt{rel-trial} needs a second-stage Griffin--TabPFN removal test for high-ranked unflagged columns such as the missed numerical proxy.

\begin{table}[H]
\centering
\caption{Per-task prediction impact and post-detection removal. Full denotes the 20-column stress test. The 0-hop, 1-hop, and 2-hop rows contain the three matched proxy designs. \(\Delta_{\mathrm{L-C}}\) is Leaky minus Clean, \(\Delta_{\mathrm{P-L}}\) is Flagged minus Leaky, and \(\Delta_{\mathrm{P-C}}\) is Flagged minus Clean. Active is the number of injected leakers present, and Removed is the total number of flagged columns removed, including false positives.}
\label{tab:task_prediction}
\scriptsize
\setlength{\tabcolsep}{2.5pt}
\resizebox{\textwidth}{!}{
\begin{tabular}{lllcccccccc}
\toprule
\textbf{Dataset} &
\textbf{Task} &
\textbf{Hop} &
\textbf{Clean F1.} &
\textbf{Leaky F1.} &
\textbf{Flagged F1.} &
\(\boldsymbol{\Delta_{\mathrm{L-C}}}\) &
\(\boldsymbol{\Delta_{\mathrm{P-L}}}\) &
\(\boldsymbol{\Delta_{\mathrm{P-C}}}\) &
\textbf{Active} &
\textbf{Removed} \\
\midrule

\texttt{rel-avito} & \texttt{user-clicks} & Clean &
0.496 & 0.496 & 0.496 &
0.000 & 0.000 & 0.000 & 0 & 0 \\
\texttt{rel-avito} & \texttt{user-clicks} & Full &
0.496 & 0.496 & 0.496 &
0.000 & 0.000 & 0.000 & 20 & 0 \\
\texttt{rel-avito} & \texttt{user-clicks} & 0-hop &
0.496 & 0.496 & 0.496 &
0.000 & 0.000 & 0.000 & 3 & 0 \\
\texttt{rel-avito} & \texttt{user-clicks} & 1-hop &
0.496 & 0.496 & 0.496 &
0.000 & 0.000 & 0.000 & 3 & 0 \\
\texttt{rel-avito} & \texttt{user-clicks} & 2-hop &
0.496 & 0.496 & 0.496 &
0.000 & 0.000 & 0.000 & 3 & 0 \\
\midrule

\texttt{rel-avito} & \texttt{user-visits} & Clean &
0.460 & 0.460 & 0.460 &
0.000 & 0.000 & 0.000 & 0 & 0 \\
\texttt{rel-avito} & \texttt{user-visits} & Full &
0.460 & 0.460 & 0.460 &
0.000 & 0.000 & 0.000 & 20 & 0 \\
\texttt{rel-avito} & \texttt{user-visits} & 0-hop &
0.460 & 0.460 & 0.460 &
0.000 & 0.000 & 0.000 & 3 & 0 \\
\texttt{rel-avito} & \texttt{user-visits} & 1-hop &
0.460 & 0.460 & 0.460 &
0.000 & 0.000 & 0.000 & 3 & 0 \\
\texttt{rel-avito} & \texttt{user-visits} & 2-hop &
0.460 & 0.460 & 0.460 &
0.000 & 0.000 & 0.000 & 3 & 0 \\
\midrule

\texttt{rel-event} & \texttt{user-ignore} & Clean &
0.722 & 0.722 & 0.622 &
0.000 & -0.100 & -0.100 & 0 & 17 \\
\texttt{rel-event} & \texttt{user-ignore} & Full &
0.722 & 0.470 & 0.470 &
-0.252 & 0.000 & -0.252 & 20 & 23 \\
\texttt{rel-event} & \texttt{user-ignore} & 0-hop &
0.722 & 0.470 & 0.612 &
-0.252 & +0.142 & -0.110 & 3 & 19 \\
\texttt{rel-event} & \texttt{user-ignore} & 1-hop &
0.722 & 0.722 & 0.599 &
+0.001 & -0.123 & -0.123 & 3 & 19 \\
\texttt{rel-event} & \texttt{user-ignore} & 2-hop &
0.722 & 0.728 & 0.489 &
+0.006 & -0.239 & -0.233 & 3 & 17 \\
\midrule

\texttt{rel-event} & \texttt{user-repeat} & Clean &
0.626 & 0.626 & 0.552 &
0.000 & -0.073 & -0.073 & 0 & 17 \\
\texttt{rel-event} & \texttt{user-repeat} & Full &
0.626 & 0.564 & 0.622 &
-0.061 & +0.058 & -0.003 & 20 & 24 \\
\texttt{rel-event} & \texttt{user-repeat} & 0-hop &
0.626 & 0.530 & 0.566 &
-0.096 & +0.036 & -0.059 & 3 & 19 \\
\texttt{rel-event} & \texttt{user-repeat} & 1-hop &
0.626 & 0.654 & 0.523 &
+0.028 & -0.131 & -0.103 & 3 & 20 \\
\texttt{rel-event} & \texttt{user-repeat} & 2-hop &
0.626 & 0.655 & 0.567 &
+0.029 & -0.088 & -0.059 & 3 & 17 \\
\midrule

\texttt{rel-trial} & \texttt{study-outcome} & Clean &
0.500 & 0.500 & 0.486 &
0.000 & -0.014 & -0.014 & 0 & 1 \\
\texttt{rel-trial} & \texttt{study-outcome} & Full &
0.500 & 0.498 & 0.443 &
-0.003 & -0.054 & -0.057 & 20 & 5 \\
\texttt{rel-trial} & \texttt{study-outcome} & 0-hop &
0.500 & 0.435 & 0.434 &
-0.065 & -0.001 & -0.066 & 3 & 2 \\
\texttt{rel-trial} & \texttt{study-outcome} & 1-hop &
0.500 & 0.542 & 0.525 &
+0.041 & -0.016 & +0.025 & 3 & 0 \\
\texttt{rel-trial} & \texttt{study-outcome} & 2-hop &
0.500 & 0.484 & 0.490 &
-0.016 & +0.006 & -0.010 & 3 & 0 \\

\bottomrule
\end{tabular}
}
\end{table}

\begin{table}[H]
\centering
\caption{Per-task IG detection results. AUPRC and AUROC are computed from the continuous attribution scores, while precision, recall, and detection F1 use the heuristic cutoff. Col. is the number of candidate columns and Flagged
is the number passing the threshold. For leakage rows with no flagged columns, precision, recall, and detection F1 are reported as zero.}
\label{tab:task_detection}
\scriptsize
\setlength{\tabcolsep}{2.3pt}
\resizebox{\textwidth}{!}{
\begin{tabular}{lllrrrrrrrrrrr}
\toprule
\textbf{Dataset} &
\textbf{Task} &
\textbf{Hop} &
\textbf{TP} &
\textbf{FP} &
\textbf{FN} &
\textbf{TN} &
\textbf{Prec.} &
\textbf{Rec.} &
\textbf{F1} &
\textbf{AUPRC} &
\textbf{AUROC} &
\textbf{Col.} &
\textbf{Flagged} \\
\midrule

\texttt{rel-avito} & \texttt{user-clicks} & Full &
0 & 0 & 20 & 30 & 0.000 & 0.000 & 0.000 & 0.803 & 0.823 & 50 & 0 \\
\texttt{rel-avito} & \texttt{user-clicks} & 0-hop &
0 & 0 & 3 & 30 & 0.000 & 0.000 & 0.000 & 0.587 & 0.911 & 33 & 0 \\
\texttt{rel-avito} & \texttt{user-clicks} & 1-hop &
0 & 0 & 3 & 30 & 0.000 & 0.000 & 0.000 & 0.175 & 0.700 & 33 & 0 \\
\texttt{rel-avito} & \texttt{user-clicks} & 2-hop &
0 & 0 & 3 & 30 & 0.000 & 0.000 & 0.000 & 0.101 & 0.422 & 33 & 0 \\
\midrule

\texttt{rel-avito} & \texttt{user-visits} & Full &
0 & 0 & 20 & 30 & 0.000 & 0.000 & 0.000 & 0.705 & 0.807 & 50 & 0 \\
\texttt{rel-avito} & \texttt{user-visits} & 0-hop &
0 & 0 & 3 & 30 & 0.000 & 0.000 & 0.000 & 0.587 & 0.911 & 33 & 0 \\
\texttt{rel-avito} & \texttt{user-visits} & 1-hop &
0 & 0 & 3 & 30 & 0.000 & 0.000 & 0.000 & 0.181 & 0.711 & 33 & 0 \\
\texttt{rel-avito} & \texttt{user-visits} & 2-hop &
0 & 0 & 3 & 30 & 0.000 & 0.000 & 0.000 & 0.097 & 0.400 & 33 & 0 \\
\midrule

\texttt{rel-event} & \texttt{user-ignore} & Full &
9 & 14 & 11 & 111 & 0.391 & 0.450 & 0.419 & 0.781 & 0.911 & 145 & 23 \\
\texttt{rel-event} & \texttt{user-ignore} & 0-hop &
3 & 16 & 0 & 109 & 0.158 & 1.000 & 0.273 & 0.667 & 0.987 & 128 & 19 \\
\texttt{rel-event} & \texttt{user-ignore} & 1-hop &
3 & 16 & 0 & 109 & 0.158 & 1.000 & 0.273 & 0.116 & 0.880 & 128 & 19 \\
\texttt{rel-event} & \texttt{user-ignore} & 2-hop &
0 & 17 & 3 & 108 & 0.000 & 0.000 & 0.000 & 0.041 & 0.531 & 128 & 17 \\
\midrule

\texttt{rel-event} & \texttt{user-repeat} & Full &
10 & 14 & 10 & 111 & 0.417 & 0.500 & 0.455 & 0.856 & 0.963 & 145 & 24 \\
\texttt{rel-event} & \texttt{user-repeat} & 0-hop &
3 & 16 & 0 & 109 & 0.158 & 1.000 & 0.273 & 1.000 & 1.000 & 128 & 19 \\
\texttt{rel-event} & \texttt{user-repeat} & 1-hop &
3 & 17 & 0 & 108 & 0.150 & 1.000 & 0.261 & 0.116 & 0.880 & 128 & 20 \\
\texttt{rel-event} & \texttt{user-repeat} & 2-hop &
1 & 16 & 2 & 109 & 0.059 & 0.333 & 0.100 & 0.073 & 0.776 & 128 & 17 \\
\midrule

\texttt{rel-trial} & \texttt{study-outcome} & Full &
5 & 0 & 15 & 117 & 1.000 & 0.250 & 0.400 & 0.834 & 0.906 & 137 & 5 \\
\texttt{rel-trial} & \texttt{study-outcome} & 0-hop &
2 & 0 & 1 & 117 & 1.000 & 0.667 & 0.800 & 0.833 & 0.991 & 120 & 2 \\
\texttt{rel-trial} & \texttt{study-outcome} & 1-hop &
0 & 0 & 3 & 117 & 0.000 & 0.000 & 0.000 & 0.533 & 0.986 & 120 & 0 \\
\texttt{rel-trial} & \texttt{study-outcome} & 2-hop &
0 & 0 & 3 & 117 & 0.000 & 0.000 & 0.000 & 0.029 & 0.430 & 120 & 0 \\

\bottomrule
\end{tabular}
}
\end{table}

\section{Per-Column Attribution Rankings and Thresholding}
\label{app:attribution_thresholding}

The detection results above depend on both the attribution ranking and the threshold applied to that ranking. We compute IG through the frozen Griffin encoder on the validation batches. Candidate representations are interpolated from a zero-feature baseline using $12$ midpoint steps, with the sampling seed reset so that the relational neighborhood and support examples remain fixed along the path. The attribution objective is the margin between the Ridge surrogate's full-input predicted class and its strongest alternative. Absolute IG values are averaged across embedding dimensions, then aggregated by column using the overall mean and the mean of the top 10\% of occurrences within each batch, combined by their geometric mean.

After a log transform, each column is standardized within its source-table and feature-family group using the median and scaled MAD; groups with fewer than ten columns use global statistics. For robust score \(z_i\), we compute the one-sided normal-tail score \(t_i = 1 - \Phi(z_i)\) and flag the column when \(m t_i < 0.05\), where \(m\) is the total number of scored columns in the run. This is numerically equivalent to the cutoff \(z_i > \Phi^{-1}(1 - 0.05/m)\) under a standard-normal reference distribution.

The \(t_i\) values are score transforms rather than calibrated null p-values, so this cutoff does not provide formal family-wise error control. In the clean \texttt{rel-event} runs, 17 original columns cross the cutoff, indicating that the present threshold is not sufficiently calibrated for automatic removal. We therefore treat the continuous IG scores as a ranking signal and the thresholded rule as an initial heuristic.

This explains why ranking and thresholded detection can differ. On \texttt{rel-avito}, the matched zero-hop leakers rank above many original columns but do not cross the heuristic cutoff. On \texttt{rel-event}, all three matched zero-hop leakers cross it, but 16 original columns do as well. On the matched zero-hop \texttt{rel-trial} run, the two text leakers are flagged while the high-ranked numerical proxy remains below the cutoff. These results support using IG for screening while treating automatic removal as a separate calibration problem.

\section{Bimodal Encoder Robustness Check}
\label{app:bimodal}

To test encoder dependence, we repeat the leakage-impact ablations with a Bimodal relational encoder while keeping the TabPFN-based ICL head and downstream evaluation protocol unchanged. Bimodal replaces Griffin's gated relation-wise mean/max message passing with a dual-attention block combining relation-conditioned local attention and type-agnostic global attention, followed by learned fusion. Across 13 tasks, mean $\Delta F1$ is $-0.115$, $-0.002$, and $+0.001$ for the matched zero-, one-, and two-hop conditions, respectively. In this additional encoder check, the mean effect is again larger at zero-hop than at one- or two-hops, while individual task effects remain variable.

\begin{table}[t]
\centering
\scriptsize
\caption{Bimodal encoder robustness check. $\Delta F1$ is relative to the corresponding clean-support baseline.}
\begin{tabular}{llrrrrr}
\toprule
Dataset & Task & Clean & 0-hop & 1-hop & 2-hop & Full \\
\midrule
rel-arxiv & paper-citation & 0.675 & -0.306 & +0.011 & +0.015 & -0.314 \\
rel-avito & searchinfo-isuserloggedon & 0.412 & +0.085 & 0.000 & 0.000 & +0.050 \\
rel-avito & searchstream-click & 1.000 & -0.870 & 0.000 & 0.000 & -0.979 \\
rel-avito & user-clicks & 0.496 & 0.000 & 0.000 & 0.000 & 0.000 \\
rel-avito & user-visits & 0.470 & 0.000 & 0.000 & 0.000 & 0.000 \\
rel-event & event\_interest-interested & 0.437 & +0.020 & 0.000 & 0.000 & +0.020 \\
rel-event & event\_interest-not\_interested & 0.498 & -0.016 & 0.000 & 0.000 & -0.085 \\
rel-event & user-ignore & 0.666 & -0.190 & -0.032 & -0.004 & -0.190 \\
rel-event & user-repeat & 0.615 & +0.010 & +0.055 & +0.003 & -0.014 \\
rel-trial & eligibilities-adult & 0.483 & 0.000 & 0.000 & 0.000 & -0.407 \\
rel-trial & eligibilities-child & 0.458 & -0.325 & 0.000 & 0.000 & -0.325 \\
rel-trial & studies-has\_dmc & 0.512 & -0.049 & -0.026 & -0.026 & -0.279 \\
rel-trial & study-outcome & 0.437 & +0.149 & -0.029 & +0.026 & 0.000 \\
\midrule
\multicolumn{3}{l}{Mean $\Delta F1$} &
-0.115 & -0.002 & +0.001 & -0.194 \\
\bottomrule
\end{tabular}
\end{table}

\section{Limitations and Future Work}
\label{app:limitations}

The primary study uses synthetic leakers, one TabPFN-based head, and five binary tasks; Appendix \ref{app:bimodal} extends the encoder and task coverage with a 13-task Bimodal robustness check. Our future work will evaluate naturally occurring leakage in addition to synthetic leakers, additional RFM architectures, and different neighborhood-sampling settings. Pretraining on a larger public relational corpus such as THE JOIN \cite{ranjanlarge} or on synthetic relational databases generated with PluRel \cite{kothapalli2026plurel} could test whether the observed leakage patterns persist under different pretraining data. Evaluation will also be extended to RelBench v2 \cite{gu2026relbench}, including its larger databases and autocomplete tasks, as well as multiclass and regression tasks.

Higher-hop leakage may vary with relational coverage, aggregation, and neighborhood sampling. Detection could be improved by validating the Ridge ranking against direct Griffin--TabPFN removal tests, testing methods that attribute the downstream ICL prediction more directly, and improving threshold calibration.

These extensions build directly on the controlled framework established here and provide a path toward more general and reliable leakage safeguards for relational ICL.

\end{document}